\documentclass{article}

\usepackage{microtype}
\usepackage{graphicx}
\usepackage{subfigure}
\usepackage{booktabs} 

\usepackage{hyperref}
\usepackage{multirow}

\usepackage[preprint]{icml2024}

\usepackage{amsmath}
\usepackage{amssymb}
\usepackage{mathtools}
\usepackage{amsthm}
\usepackage{cuted}
\usepackage[capitalize,noabbrev]{cleveref}

\theoremstyle{plain}

\theoremstyle{definition}

\theoremstyle{remark}

\setcitestyle{numbers,citesep={,},square}

\definecolor{citecolor}{HTML}{0071BC}
\definecolor{linkcolor}{HTML}{ED1C24}
\hypersetup{
    linkcolor=linkcolor,
    citecolor=citecolor,
    urlcolor=citecolor
}
\definecolor{motion-editing}{HTML}{F0A780}
\definecolor{appearance-editing}{HTML}{156082}
\definecolor{main-path}{HTML}{F2AA84}
\definecolor{recons-branch}{HTML}{A5C496}
\definecolor{motion-branch}{HTML}{8891DB}
\usepackage[textsize=tiny]{todonotes}
\usepackage{graphicx, amsmath, amssymb, caption, subcaption, multirow, overpic, textpos}

\icmltitlerunning{UniEdit: A Unified Tuning-Free Framework for Video Motion and Appearance Editing}

\begin{document}

\twocolumn[
\icmltitle{UniEdit: A Unified Tuning-Free Framework for\\Video Motion and Appearance Editing}


\icmlsetsymbol{equal}{*}
\icmlsetsymbol{corres}{$\dagger$}

\begin{icmlauthorlist}
\icmlauthor{Jianhong Bai}{equal,Zhejiang University}
\icmlauthor{Tianyu He}{corres,Microsoft Research Asia}
\icmlauthor{Yuchi Wang}{Peking University}
\icmlauthor{Junliang Guo}{Microsoft Research Asia}
\icmlauthor{Haoji Hu}{Zhejiang University}
\icmlauthor{Zuozhu Liu}{Zhejiang University}
\icmlauthor{Jiang Bian}{Microsoft Research Asia}
\end{icmlauthorlist}
\begin{center}
    {Project webpage:} \url{https://jianhongbai.github.io/UniEdit/}
    \vspace{-10pt}
\end{center}
\vspace{-60pt}
\icmlaffiliation{Zhejiang University}{Zhejiang University}
\icmlaffiliation{Microsoft Research Asia}{Microsoft Research Asia}
\icmlaffiliation{Peking University}{Peking University}


\icmlkeywords{Machine Learning, ICML, Diffusion Model, Video Editing}

\vskip 0.3in
]



\printAffiliationsAndNotice{\icmlEqualContribution\icmlcoress} 
\begin{strip}\centering

\includegraphics[width=0.95\textwidth]{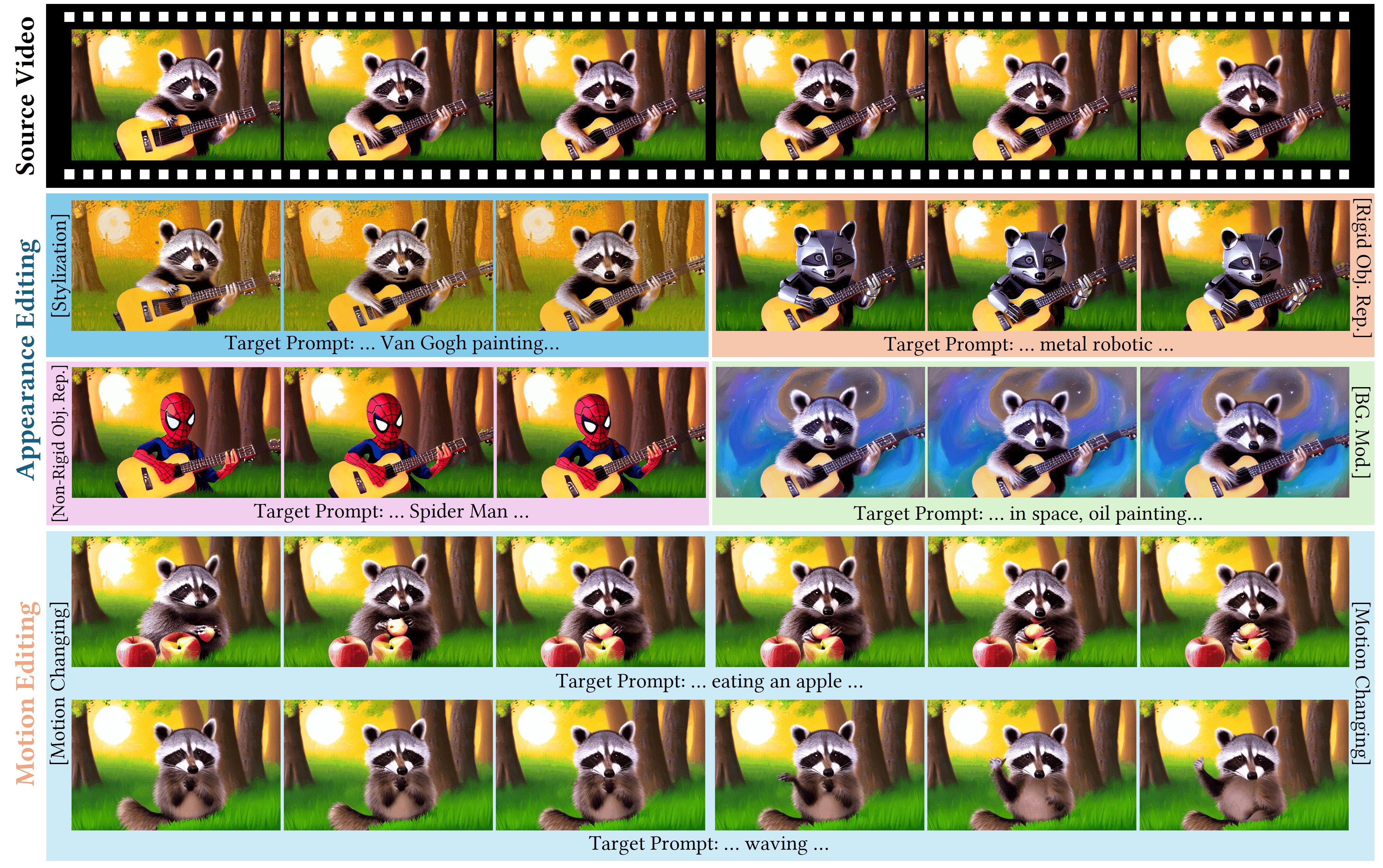}
\vspace{-10pt}
\captionof{figure}{Examples edited by UniEdit. Our solution supports both video \textit{\textbf{\textcolor{motion-editing}{motion}}} editing in the time axis (i.e., from playing guitar to eating or waving) and various video \textit{\textbf{\textcolor{appearance-editing}{appearance}}} editing scenarios (i.e., stylization, rigid/non-rigid object replacement, background modification). We encourage the readers to watch the videos on our \href{https://jianhongbai.github.io/UniEdit/}{project page}.}
\label{fig_1}
\end{strip}

\begin{abstract}

Recent advances in text-guided video editing have showcased promising results in appearance editing (e.g., stylization). However, video motion editing in the temporal dimension (e.g., from eating to waving), which distinguishes video editing from image editing, is underexplored. In this work, we present UniEdit, a tuning-free framework that supports both video motion and appearance editing by harnessing the power of a pre-trained text-to-video generator within an inversion-then-generation framework.
To realize motion editing while preserving source video content, based on the insights that temporal and spatial self-attention layers encode inter-frame and intra-frame dependency respectively, we introduce auxiliary motion-reference and reconstruction branches to produce text-guided motion and source features respectively. The obtained features are then injected into the main editing path via temporal and spatial self-attention layers.
Extensive experiments demonstrate that UniEdit covers video motion editing and various appearance editing scenarios, and surpasses the state-of-the-art methods. Our code will be publicly available.

\end{abstract}

\section{Introduction}

The advent of pre-trained diffusion-based~\citep{ddpm,ddim} text-to-image generators~\citep{sd,imagen,dalle2} has revolutionized the fields of design and filmmaking, opening new vistas for creative expression. These advancements, underpinned by seminal works in text-to-image synthesis, have paved the way for innovative text-guided editing techniques for both images~\citep{sdedit,p2p,brooks2023instructpix2pix,masactrl} and videos~\citep{tune-a-video,pix2vid,videop2p,rerender,ccedit,fatezero}. Such techniques not only enhance creative workflows but also promise to redefine content creation within these industries.

Video editing, in contrast to image editing, introduces the intricate challenge of ensuring frame-wise consistency. Efforts to address this challenge have led to the development of methods that leverage shared features and structures with the source video~\citep{pix2vid,videop2p,t2v_zero,rerender,fatezero,stablevideo,ground-a-video,wang2023zero,tokenflow} through an inversion-then-generation pipeline~\citep{sdedit,ddim}, exemplified by Pix2Video's approach~\citep{pix2vid} to consistent appearance editing across frames. To transfer the edited appearance from the anchor frame to the remaining frames consistently, it employs a pre-trained image generator and extends the self-attention layers to cross-frame attention to generate each remaining frame. Despite these advancements in performing video \textit{appearance} editing (e.g., stylization, object appearance replacement, etc.), these methodologies fall short in editing video \textit{motion} (e.g., replacing the movement of playing guitar with waving), hampered by a lack of motion priors and limited control over inter-frame dependencies, underscoring a critical gap in video editing capabilities.

Previous attempts~\cite{tune-a-video,dreamix} at video motion editing through fine-tuning a pre-trained generator on the given source video and then editing motion through text guidance. Although effective, they necessitate a delicate balance between the generative prowess of the model and the preservation of the source video's content. This compromise often leads to restricted motion diversity and unwanted content variations, indicating a pressing need for a more robust solution.

In response, our work aims to explore a \textit{tuning-free} framework that adeptly navigates the complexities of editing both the \textit{motion} and \textit{appearance} of videos. To achieve this, we identify three technical challenges: 1) it is non-trivial to incorporate the text-guided motion into the source content, as directly applying video appearance editing~\citep{fatezero,tokenflow} or image editing~\citep{masactrl} schemes leads to undesirable results (as shown in Fig.~\ref{fig_exp_comparison}); 2) non-edited content preservation of the source video; 3) inheriting the spatial structure of the source video during appearance editing.

Our solution, UniEdit, harnesses the power of a pre-trained text-to-video generator (e.g., LaVie~\citep{lavie}) within an inversion-then-generation framework~\citep{sdedit}, tailored to overcome the identified challenges. Particularly, we introduce three key innovations: 1) To inject text-guided motion into the source content, we highlight the insight that \textit{the temporal self-attention layers of the generator encode the inter-frame dependency}. Acting in this way, we introduce an auxiliary motion-reference branch to generate text-guided motion features, which are then injected into the main editing path via temporal self-attention layers. 2) To preserve the non-edited content of the source video, motivated by the image editing technique~\citep{masactrl}, we follow the insight that \textit{the spatial self-attention layers of the generator encode the intra-frame dependency}. Therefore, we introduce an auxiliary reconstruction branch, and inject the features obtained from the spatial self-attention layers of the reconstruction branch into the main editing path. 3) To retain the spatial structure during the appearance editing, we replace the spatial attention maps of the main editing path with those in the reconstruction branch. These innovations collectively enable UniEdit to redefine the landscape of video editing.

To our best knowledge, UniEdit represents a pioneering leap in text-guided, tuning-free video motion editing. In addition, its unified architecture not only facilitates a wide array of video appearance editing tasks, as shown in Fig.~\ref{fig_1}, but also empowers image-to-video generators for zero-shot text-image-to-video generation. Through comprehensive experimentation, we demonstrate UniEdit's superior performance relative to existing state-of-the-art methods, highlighting its potential to significantly advance the field of video editing.

\vspace{-2mm}
\section{Related Works}
\vspace{-1mm}
\subsection{Video Generation}

Researchers have achieved video generation with generative adversarial networks~\citep{vgan,tgan,vid2vid}, language models~\citep{videogpt,magvit}, or diffusion models~\citep{vdm,make-a-video,imagenvideo,he2022latent,blattmann2023align,modelscope,show1,girdhar2023emu,lavie,videocrafter1}. To make the generation more controllable, recent endeavors have also incorporated additional structure guidance (e.g., depth map)~\citep{gen1,chen2023motion,controlvideo,control-a-video,sparsectrl,motionctrl}, or conducted customized generation~\citep{tune-a-video,make-your-video,videobooth,motioncrafter,dreamvideo,materzynska2023customizing}. These models have generally learned real-world video distribution from large-scale data, and achieved promising results on text-to-video or image-to-video generation. Based on their success, we leverage the learned prior in the pre-trained model to achieve tuning-free video motion and appearance editing.

\vspace{-2mm}
\subsection{Video Editing}

Video editing aims to produce a new video that is aligned with the provided editing instructions (e.g., text) while maintaining the other characteristics of the source video. It can be categorized into appearance and motion editing.

For appearance editing~\citep{rerender, dragvideo, ccedit, kara2023rave, cong2023flatten}, like turn a video into the style of Van Gogh, the main challenge is to achieve temporal-consistent generation across different frames. Early attempts~\citep{pix2vid, t2v_zero, fatezero, stablevideo, ground-a-video, wang2023zero} leveraged text-to-image models with inter-frame propagation to ensure consistency. For instance, Pix2Video~\cite{pix2vid} replaces the key and value of the current frame with those of the first and previous frame. Video-P2P~\citep{videop2p} achieved local editing via video-specific fine-tuning and unconditional embedding optimization~\citep{null_text_inversion}. Follow-up studies~\citep{tokenflow,rerender,codef} also leveraged the edit-then-propagate framework with neatest-neighbor field~\citep{tokenflow}, estimated optical flow~\citep{rerender}, or temporal deformation field~\citep{codef}. Despite the promising results, due to the constraint on the source video structure, these approaches are specialized in appearance editing and can not be applied to motion editing directly.

Recent studies have also explored video motion editing with text guidance~\citep{tune-a-video,dreamix}, user-provided motion~\citep{vmc,drag-a-video,dragvideo}, or specific motion representation~\citep{tu2023motioneditor,karras2023dreampose,he2023gaia}. For example, Dreamix~\cite{dreamix} proposed fine-tuning a pre-trained text-to-video model with mixed video-image reconstruction objectives for each source video. Then the editing is realized by conditioning the fine-tuned model on the given target prompt. MoCA~\citep{yan2023motion} decoupled the video into the first-frame appearance and the optical flow, and trained a diffusion model to generate video conditioned on the first frame and the text. However, it struggled to preserve the non-edited motion (e.g., background dynamics) as it generates the entire motion from the text.
Different from the aforementioned approaches that require fine-tuning or user-provided motion input, we are the first to achieve tuning-free motion and appearance editing with text guidance only.

\vspace{-2mm}
\section{Preliminaries: Video Diffusion Models}

Our proposed UniEdit is built upon video diffusion models. Therefore, we first recap the architecture that is used in common text-guided video diffusion models~\citep{lavie,svd}.

\vspace{-0mm}
\subsection{Overall Architecture}
\label{sec_pre_arch}

Modern text-to-video (T2V) diffusion models typically extend a pre-trained text-to-image (T2I) model~\citep{sd} to the video domain with the following adaptations. 1) Introducing additional temporal layers by inflating 2d convolutional layers to 3d form, or adding temporal self-attention layers~\citep{vaswani2017attention} to model the correlation between video frames. 2) Due to the extensive computational resources for modeling spatial-temporal joint distribution, these works typically first train video generation models on low spatial and temporal resolutions, and then upsampling the generated results with cascaded models. 3) Other improvements like efficiency~\citep{bar2024lumiere}, training strategy~\citep{girdhar2023emu}, or additional control signals~\citep{gen1}, etc.

During inference, given standard Gaussian distribution $z_T \sim \mathcal{N}(0, 1)$, the trained denoising UNet is used to perform $T$ denoising steps to obtain the outputs~\citep{ddpm,ddim}. If the model is trained in latent space~\citep{sd}, a decoder is employed to reconstruct videos from the latent domain.

\vspace{-2mm}
\subsection{Attention Mechanisms}

In particular, for each block of the denoising UNet, there are four basic modules: a convolutional module, a spatial self-attention module (SA-S), a spatial cross-attention module (CA-S), and a temporal self-attention module (SA-T). Formally, the attention operation~\citep{vaswani2017attention} can be formulated as:
\begin{equation}
    \texttt{attn}(Q, K, V) = \texttt{softmax}(\frac{Q K^T}{\sqrt{d}}) V,
\label{eq_vanilla_attn}
\end{equation}
where $Q$ (query), $K$ (key), $V$ (value) are derived from the inputs, and $d$ is the dimension of the hidden states.

Intuitively, CA-S is in charge of fusing semantics from the text condition, SA-S models the intra-frame dependency, SA-T models the inter-frame dependency and ensures the generated results are temporally consistent. We leverage these intuitions in our designs as elaborated below.

\begin{figure*}[t]\centering
\includegraphics[width=1\textwidth]{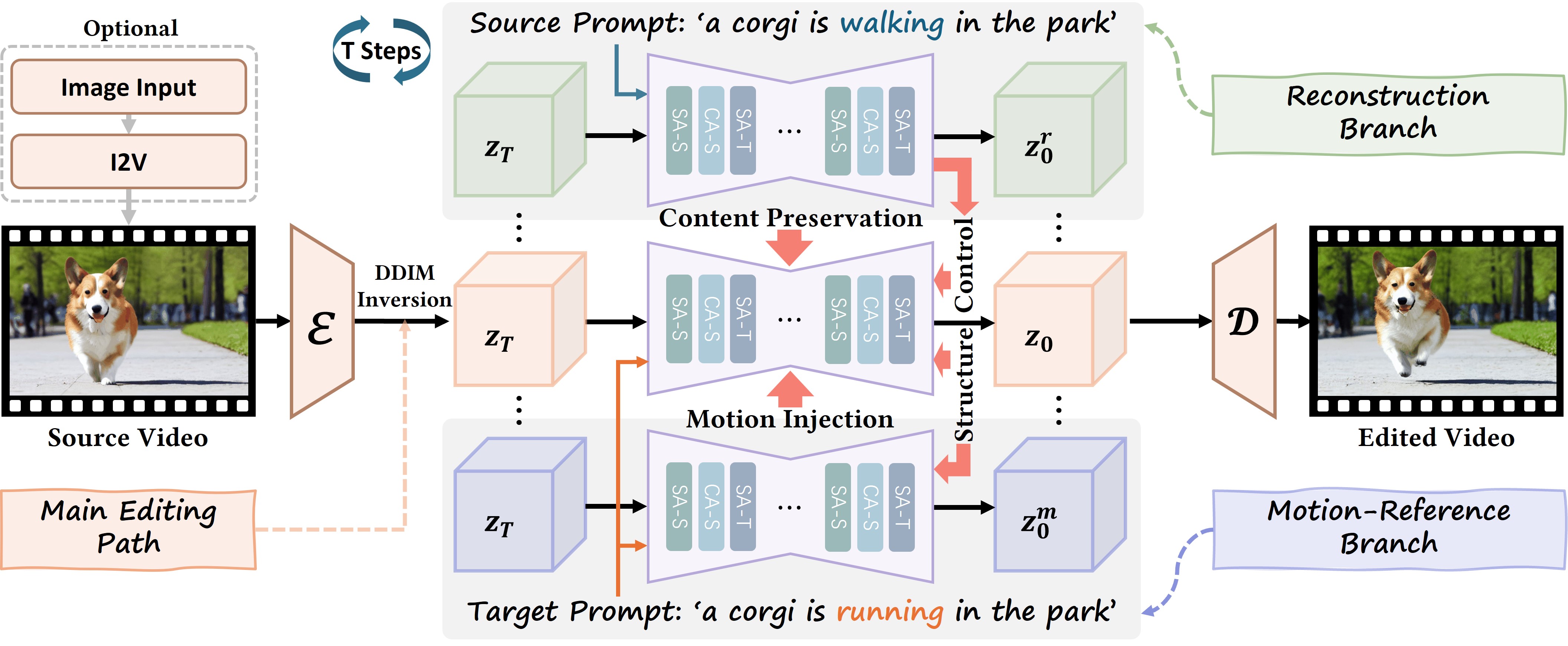}
\vspace{-5mm}
\caption{Overview of UniEdit. It follows an inversion-then-generation pipeline and consists of a \textcolor{main-path}{main editing path}, an auxiliary \textcolor{recons-branch}{reconstruction branch} and an auxiliary \textcolor{motion-branch}{motion-reference branch}. The reconstruction branch produces source features for content preservation, and the motion-reference branch yields text-guided motion features for motion injection. The source features and motion features are injected into the main editing path through spatial self-attention (SA-S) and temporal self-attention (SA-T) modules respectively (Sec.~\ref{sec_motion_editing}). We further introduce spatial structure control to retain the coarse structure of the source video (Sec.~\ref{sec_appearance_editing}).}
    \label{fig_pipe}
\vspace{-3mm}
\end{figure*}

\vspace{-2mm}
\section{UniEdit}
\label{sec_method}

\begin{figure}[t]
    \centering
    \includegraphics[width=0.48\textwidth]{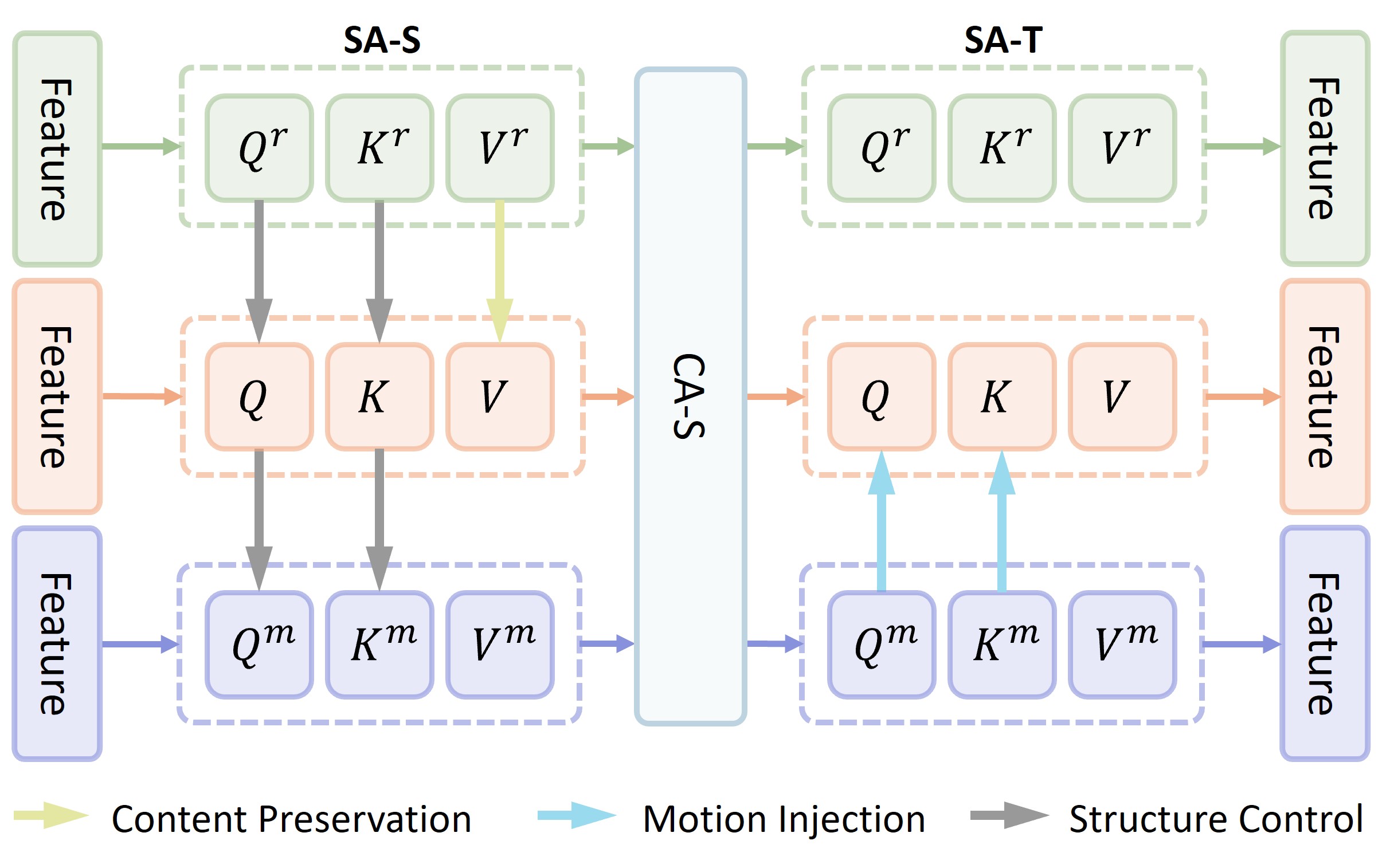}
    \vspace{-6mm}
    \caption{Detailed illustration of the relationship between the \textcolor{main-path}{main editing path}, the auxiliary \textcolor{recons-branch}{reconstruction branch} and the auxiliary \textcolor{motion-branch}{motion-reference branch}. The content preservation, motion injection and spatial structure control are achieved by the fusion of $Q$ (query), $K$ (key), $V$ (value) features in spatial self-attention (SA-S) and temporal self-attention (SA-T) modules.} 
    \label{fig_detailed_pipe}
    \vspace{-3mm}
\end{figure}

\paragraph{Method Overview.}
As shown in Fig.~\ref{fig_pipe}, our main editing path is based on an inversion-then-generation pipeline: we use the latent after DDIM inversion~\citep{ddim} as the initial noise $z_T$\footnote{For real source video, we set source prompt to null during both forward and inversion process to achieve high-quality reconstruction~\citep{null_text_inversion}.}, then perform denoising process starting from $z_T$ with the pre-trained UNet conditioned on the target prompt $P_t$. For motion editing, to achieve source content preservation and motion control, we propose to incorporate an auxiliary reconstruction branch and an auxiliary motion-reference branch to provide desired source and motion features, which are injected into the main editing path to achieve content preservation and motion editing (as shown in Fig.~\ref{fig_detailed_pipe}). We propose the pipeline of motion editing and appearance editing in Sec.~\ref{sec_motion_editing} \& Sec.~\ref{sec_appearance_editing} respectively. To further alleviate the background inconsistency, we introduce a mask-guided coordination scheme in Sec.~\ref{sec_maskedattn_ctrl}. We also extend UniEdit to text-image-to-video generation (TI2V) in Sec.~\ref{sec_ti2v}.

\vspace{-1mm}
\subsection{Tuning-Free Video Motion Editing}
\label{sec_motion_editing}

\paragraph{Content Preservation on SA-S Modules.}
One of the key challenges of editing tasks is to inherit the original content (e.g., textures and background) in the source video. To this end, we introduce an auxiliary reconstruction branch. The reconstruction path starts from the same inversed latent $z_T$ similar to the main editing path, and then conducts the denoising process with the pre-trained UNet conditioned on the source prompt $P_s$ to reconstruct the original frames. As verified in image editing~\cite{tumanyan2023plug, p2p, masactrl}, the attention features in the denoising model during reconstruction contain the content of the source video. Hence, we inject attention features of the reconstruction path into the main editing path on spatial self-attention (SA-S) layers for content preservation. At denoising step $t$, the attention operation of the $l$-th SA-S module in the main editing path is formulated as:
\begin{equation}
    \text{SA-S}^{l}_{\text{edit}} := 
    \begin{cases}
        \texttt{attn}(Q, K, V^r), & t>t_0 \,\text{and}\, l>l_0 \\
        \texttt{attn}(Q, K, V), & \text{otherwise}
    \end{cases}
\label{eq_replace_sa_kv}
\end{equation}
where $Q$, $K$, $V$ are features in the main editing path, $V^r$ refer to the value feature of the corresponding SA-S layer in the reconstruction branch, $t_0$ and $l_0$ are hyper-parameters following previous work~\citep{masactrl}. By replacing the value of spatial features, the video synthesized by the main editing path retains the non-edited characters (e.g., identity and background) of the source video, as exhibited in Fig. \ref{fig_ablation1}. Unlike previous video editing works \cite{t2v_zero, free_bloom} which introduces a cross-frame attention mechanism (i.e., using the key and value of the first/last frame), we implement Eq. \ref{eq_replace_sa_kv} frame-wisely to better tackle source video with large dynamics.

\vspace{-1mm}
\paragraph{Motion Injection on SA-T Modules.} 
After implementing the content-preserving technique introduced above, we can obtain an edited video with the same content in the source video. However, it is observed that the output video could not follow the text prompt $P_t$ properly. A straightforward solution is to increase the value of $t_0$ and $l_0$ so that balancing between the impact of injected information and the conditioned text prompt. Nevertheless, this could result in a content mismatch with the original source video in terms of structures and textures.

To obtain the desired motion without sacrificing content consistency, we propose to guide the main editing path with reference motion. Concretely, an auxiliary motion-reference branch (which also starts from the inversed latent $z_T$) is involved during the denoising process. Different from the reconstruction branch, the motion-reference branch is conditioned on the target prompt $P_t$, which contains the description of the desired motion. To transfer the motion into the main editing path, our core insight here is that \textit{temporal layers model the inter-frame dependency of the synthesized video clip} (as shown in Fig.~\ref{fig_vis}). Motivated by the observations above, we design the attention map injection on temporal self-attention layers of the main editing path: 
\begin{equation}
    \text{SA-T}^{l}_{\text{edit}} := 
    \begin{cases}
    \texttt{attn}(Q^m, K^m, V), &t>t_1 \,\text{and}\, l>l_1 \\
    \texttt{attn}(Q, K, V), \quad &\text{otherwise}
    \end{cases}
\label{eq_replace_temp_qk}
\end{equation}
where $Q^m$ and $K^m$ refer to the query and key of the motion-reference branch, we simply set $t_1$ and $l_1$ to zero in practice. It's observed that the injection of temporal attention maps can effectively facilitate the main editing path to generate motion aligned with the target prompt. To better fuse the motion with the content in the source video, we also implement spatial structure control (refer to Sec. \ref{sec_appearance_editing}) on the main editing path and motion-reference branch in the early steps.

\vspace{-2mm}
\subsection{Tuning-Free Video Appearance Editing}
\label{sec_appearance_editing}
In Sec. \ref{sec_motion_editing}, we introduce the pipeline of UniEdit for video motion editing. In this subsection, we aim to perform appearance editing (e.g., style transfer, object replacement, background changing) via the same framework. 
In general, there are two main differences between appearance editing and motion editing. Firstly, appearance editing does not require changing the inter-frame relationships. Therefore, we remove the motion-reference branch and corresponding motion injection mechanism from the motion editing pipeline. Secondly, the main challenge of appearance editing is to maintain the structural consistency of the source video. To address this, we introduce spatial structure control between the main editing path and the reconstruction branch.

\vspace{-1mm}
\paragraph{Spatial Structure Control on SA-S Modules.} 
Previous approaches on video appearance editing~\citep{rerender, tokenflow} mainly realize spatial structure control with the assistance of additional network~\citep{controlnet}. When the auxiliary control model fails, it may result in inferior performance in preserving the structure of the original video. Alternatively, we suggest extracting the layout information of the source video from the reconstruction branch. Intuitively, the attention maps in spatial self-attention layers encode the structure of the synthesized video, as verified in Fig.~\ref{fig_vis}. Hence, we replace the query and key of SA-S module in the main editing path with those in the reconstruction branch:
\begin{equation}
    \text{SA-S}^{l}_{\text{edit}} := 
    \begin{cases}
    \texttt{attn}(Q^r, K^r, V), &t<t_2 \,\text{and}\, l>l_2, \\
    \texttt{attn}(Q, K, V), \; &\text{otherwise},
    \end{cases}
\label{eq_replace_sa_qk}
\end{equation}
where $Q^r$ and $K^r$ refer to the query and key of the reconstruction branch, $t_2$ and $l_2$ are used to control the extent of editing. It is worth mentioning that the effect of spatial structure control is distinct from the content preservation mechanism in Sec. \ref{sec_motion_editing}. Take stylization as an example, the proposed structure control in Eq. \ref{eq_replace_sa_qk} only ensures consistency in terms of each frame's composition, while enabling the model to generate the required textures and styles based on the text prompt. On the other hand, the content preservation technique inherits the textures and style of the source video. Therefore, we use structure control instead of content preservation for appearance editing.

In addition, the structure control technique also facilitates the fusion of source content and targeted motion in motion editing.
During motion editing, the structure control is only implemented at earlier denoising steps, and will not affect the model to generate the desired motion that matches the target prompt. 

\vspace{-1mm}
\subsection{Mask-Guided Coordination}
\label{sec_maskedattn_ctrl}

To further improve background consistency, we suggest leveraging the foreground/background segmentation mask $M$ to guide the denoising process \cite{couairon2023videdit, couairon2022diffedit}. There are two possible ways to obtain the mask $M$: the attention maps of CA-S modules with a threshold~\citep{p2p}; or employing an off-the-shelf segmentation model~\citep{sam} on the source and generated videos.
The obtained segmentation masks can be leveraged to 1), alleviate the indistinction in foreground and background; 2), improve the consistency of content between edited and source videos.

To this end, we leverage mask-guided self-attention in the main editing path to coordinate the editing path with the motion-reference branch. Formally, we define:
\begin{equation}
    \texttt{m-attn}(Q, K, V; M) = \texttt{softmax}(\frac{Q K^T}{\sqrt{d}} + M) V.
\label{eq_maskattn}
\end{equation}
Then the mask-guided self-attention:
\begin{equation}
    \begin{aligned}
    \text{SA}_{\text{mask}} := &\texttt{m-attn}(Q, K, V; M^f) \odot M_m \\
       + &\texttt{m-attn}(Q, K, V; M^b) \odot (1 - M_m),
    \end{aligned}
\label{eq_maskattn_fusion}
\end{equation}
where $M^f, M^b \in \{-\infty,0\}$ indicate the foreground and background masks in the editing path respectively, $M_m \in \{0,1\}$ denotes the foreground mask from the motion-reference branch, and $\odot$ is Hadamard product.

In addition, we leverage the mask during the content preservation and motion injection for the features obtained from the reconstruction branch and the motion-reference branch (e.g., we replace $Q^m$ with $M_m \odot Q^m + (1 - M_m) \odot Q$).

\begin{figure*}[t]\centering
\includegraphics[width=1\textwidth]{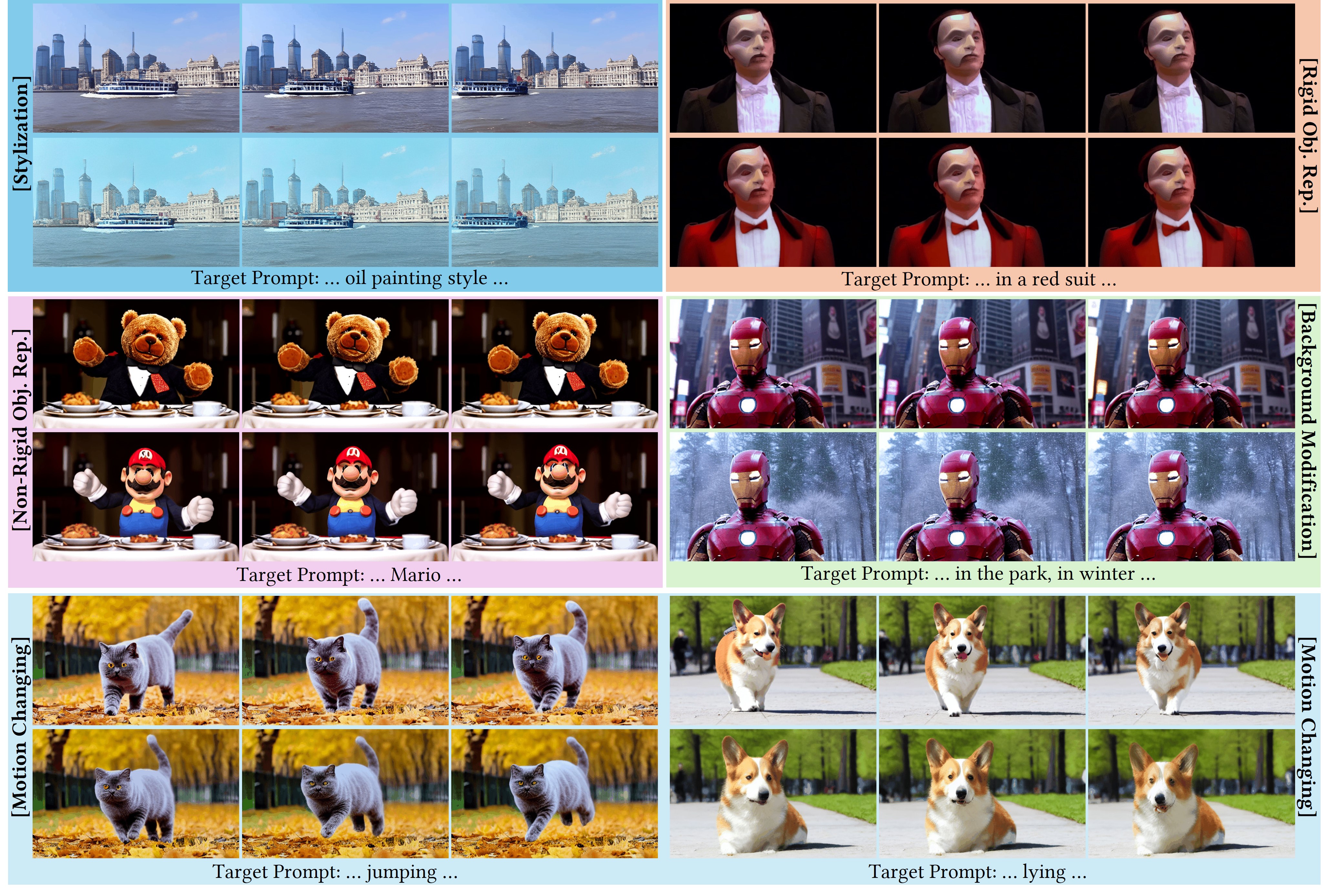}
\vspace{-6mm}
\caption{Examples edited by UniEdit. For each case, the upper frames come from the source video, and the lower frames indicate the edited results with the target prompt. We encourage the readers to watch the \href{https://jianhongbai.github.io/UniEdit/}{videos} and make evaluations.}
\label{fig_exp_ours_results}
\vspace{-3mm}
\end{figure*}

\vspace{-1mm}
\subsection{T2V Models are Zero-Shot TI2V Generators}
\label{sec_ti2v}
To make our framework more flexible, we further derive a method to incorporate images as input and synthesize high-quality video conditioned on \textit{both} image and text-prompt.
Different from some image animation techniques~\cite{svd}, our method allows the user to guide the animation process with text prompts. 
Concretely, we first achieve image-to-video (I2V) generation by: 1) transforming input images with simulated camera movement to form a pseudo-video clip~\citep{dreamix} or 2) leveraging existing image animation approaches (e.g., SVD~\cite{svd}, AnimateDiff~\cite{animatediff}) to synthesis a video with random motion (which may not consistent with the text prompt). Then, we perform text-guided editing with UniEdit on the vanilla video to obtain the final output video.

\vspace{-2mm}
\section{Experiments}

\begin{figure*}[t]\centering
\includegraphics[width=1\textwidth]{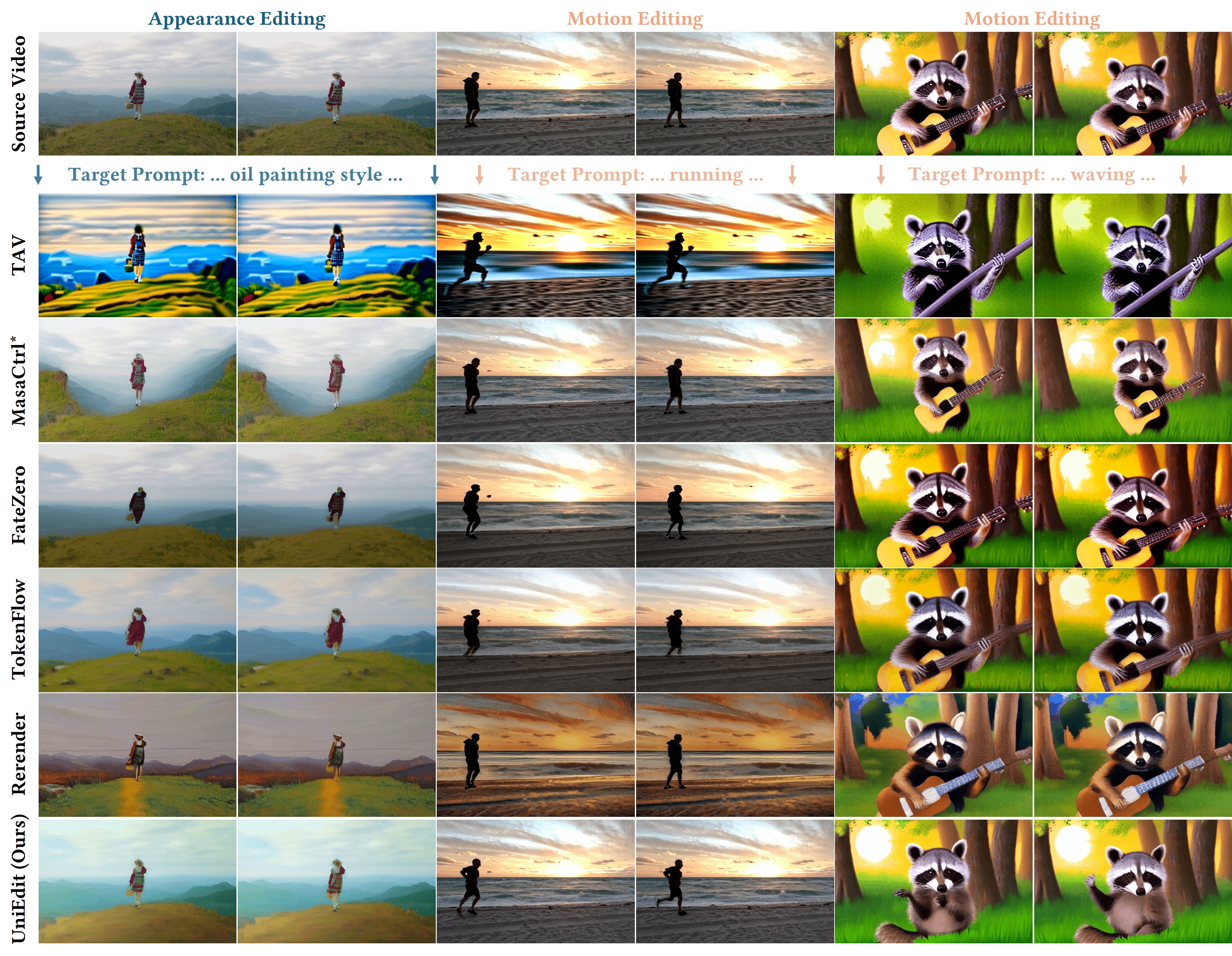}
\vspace{-6mm}
\caption{Comparison with state-of-the-art methods for both video motion and appearance editing. It shows that UniEdit achieves better source content preservation, and outperforms baselines in motion editing by a large margin.}
\label{fig_exp_comparison}
\vspace{-3mm}
\end{figure*}

\vspace{-1mm}
\subsection{Implementation Details}
UniEdit is not limited to specific video diffusion models. In this section, we build UniEdit upon LaVie~\cite{lavie} as an instantiation to verify the effectiveness of our method.
For each input video, we follow the pre-processing step in LaVie to the resolution of $320 \times 512$. Then, the pre-processed video is fed into the UniEdit to perform video editing. It takes 1-2 minutes to edit on an NVIDIA A100 GPU for each video. More details can be found in Appendix \ref{appendix_sec_B}.

\vspace{-2mm}
\subsection{Comparison with State-of-the-Art Methods}
\paragraph{Baselines.}
To evaluate the performance of UniEdit, we compare the editing results of UniEdit with state-of-the-art motion and appearance editing approaches. For motion editing, due to the lack of open-source tuning-free (zero-shot) methods, we adapt the state-of-the-art non-rigid image editing technique MasaCtrl~\cite{masactrl} to a T2V model~\cite{lavie} (denoted as MasaCtrl$^*$ in Fig. \ref{fig_exp_comparison}) and a one-shot video editing method Tune-A-Video (TAV)~\cite{tune-a-video} as strong baselines. For appearance editing, we use the latest methods with strong performance, including FateZero~\cite{fatezero}, TokenFlow~\cite{tokenflow}, and Rerender-A-Video (Rerender)~\citep{rerender} as baselines.

\vspace{-3mm}
\paragraph{Qualitative Results.}
We give editing examples of UniEdit in Fig.~\ref{fig_1}, Fig.~\ref{fig_exp_ours_results}. (more examples in Fig.~\ref{fig_appendix_ours_appear1}-\ref{fig_appendix_ours_motion2} of Appendix~\ref{appendix_sec_C}). We recommend seeing more cases and videos on our \href{https://jianhongbai.github.io/UniEdit/}{project page}. It is observed that UniEdit can: 1) edit in different scenarios, including motion-changing, object replacement, style transfer, background modification, etc; 2) align with the target prompt; 3) demonstrate great temporal consistency.

Furthermore, we make comparisons with state-of-the-art methods in Fig.~\ref{fig_exp_comparison} (more comparisons in Fig.~\ref{fig_appendix_compare1},\ref{fig_appendix_compare2} of Appendix~\ref{appendix_sec_C}). For appearance editing, i.e., turning the source video into an oil painting style, UniEdit outperforms baselines on content preservation. For example, the grassland still maintains its original appearance without any additional stones or paths. For motion editing, most baseline methods fail to output video that is aligned with the target prompt, or fail to preserve the source content.

\begin{table}[t]
        \small
	\begin{center}
		\caption{Quantitative comparison (CLIP Score and User Preference) with state-of-the-art video editing techniques.}
            \vspace{-0.3cm}
		\label{tab_quality_eval}
		\setlength\tabcolsep{2pt}
		\begin{tabular}{lcccc}
			\toprule
			\multirow{3}{*}{Method} & \multicolumn{2}{c}{Frame Consistency} & \multicolumn{2}{c}{Textual Alignment}\\ 
		      \cmidrule(r){2-3} \cmidrule(r){4-5}
                & CLIP Score & User Pref. & CLIP Score & User Pref.\\
                \midrule
                TAV~\cite{tune-a-video} & 96.88 & 3.86 & 32.93 & 3.34 \\
                MasaCtrl$^*$~\cite{masactrl} & 96.52 & 4.32 & 31.93 & 3.20 \\
                FateZero~\cite{fatezero} & 97.56 & 4.53 & 30.67 & 3.45 \\
                Rerender~\citep{rerender} & 97.17 & 4.18 & 32.18 & 3.56 \\
                TokenFlow~\cite{tokenflow} & 97.38 & 4.56 & 31.33 & 3.47 \\
                UniEdit (Ours) & \textbf{98.37} & \textbf{4.74} & \textbf{36.29} & \textbf{4.88} \\
			\bottomrule
		\end{tabular}
	\end{center}
 \vspace{-3mm}
\end{table}

\vspace{-3mm}
\paragraph{Quantitative Results.}
We quantitatively verify the effectiveness of our method from two aspects: temporal consistency and alignment with the target prompt. Following previous work~\cite{tune-a-video}, we use the CLIP~\citep{clip} to calculate the score of Frame Consistency and Textual Alignment. We also conducted a user study between UniEdit and the baselines by asking $10$ participants to rate the edited videos at five grades ($1$-$5$). As shown in Tab.~\ref{tab_quality_eval}, UniEdit outperforms baseline methods by a large margin.

\vspace{-1mm}
\subsection{Ablation Study and Analysis}

\begin{figure}[t]\centering
\includegraphics[width=0.48\textwidth]{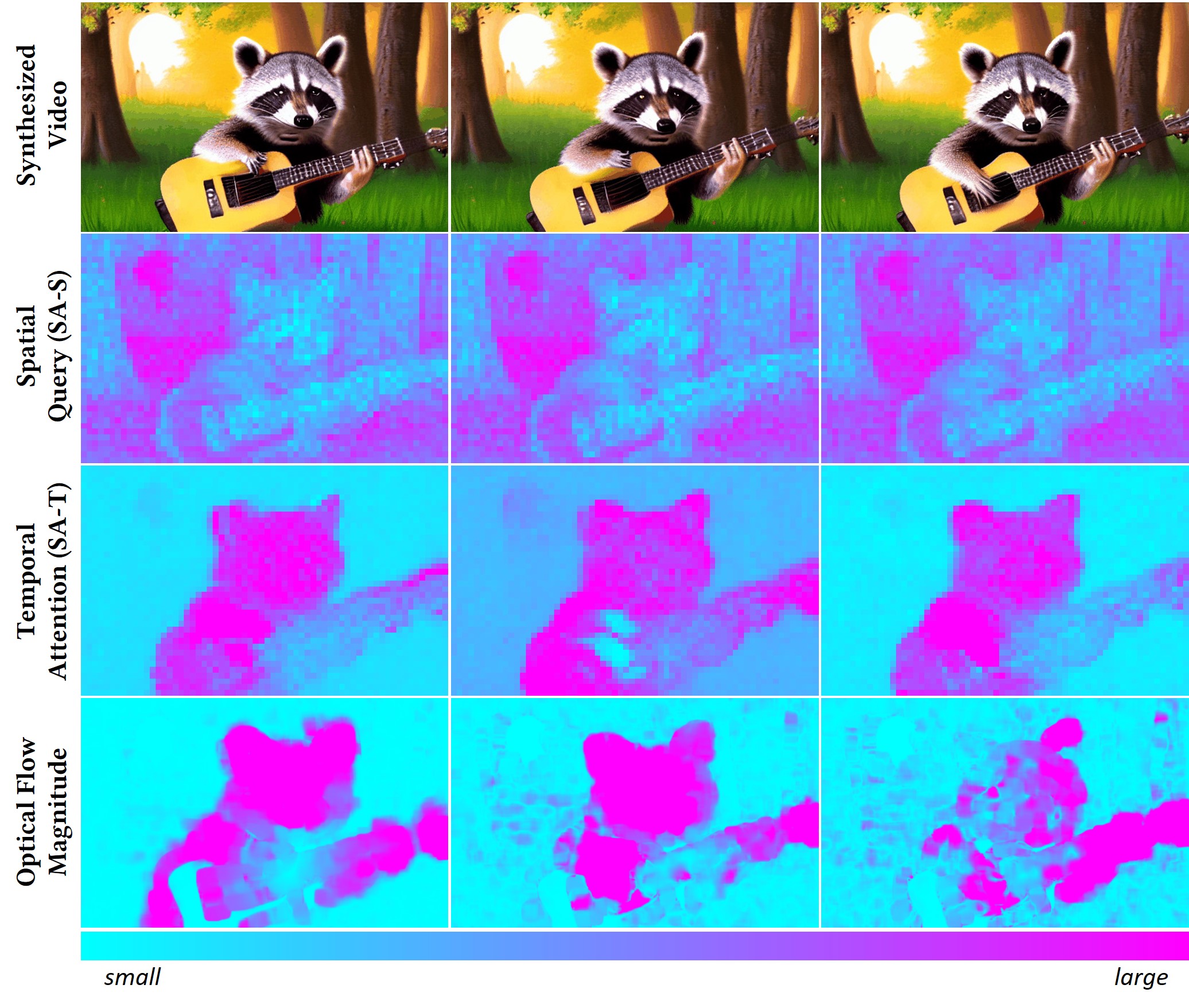}
\vspace{-8mm}
\caption{Visualization of spatial query in SA-S (second row), cross-frame temporal attention maps in SA-T (third row), and the magnitude of optical flow (fourth row).}
\label{fig_vis}
\vspace{-3mm}
\end{figure}

\begin{figure}[t]\centering
\includegraphics[width=0.48\textwidth]{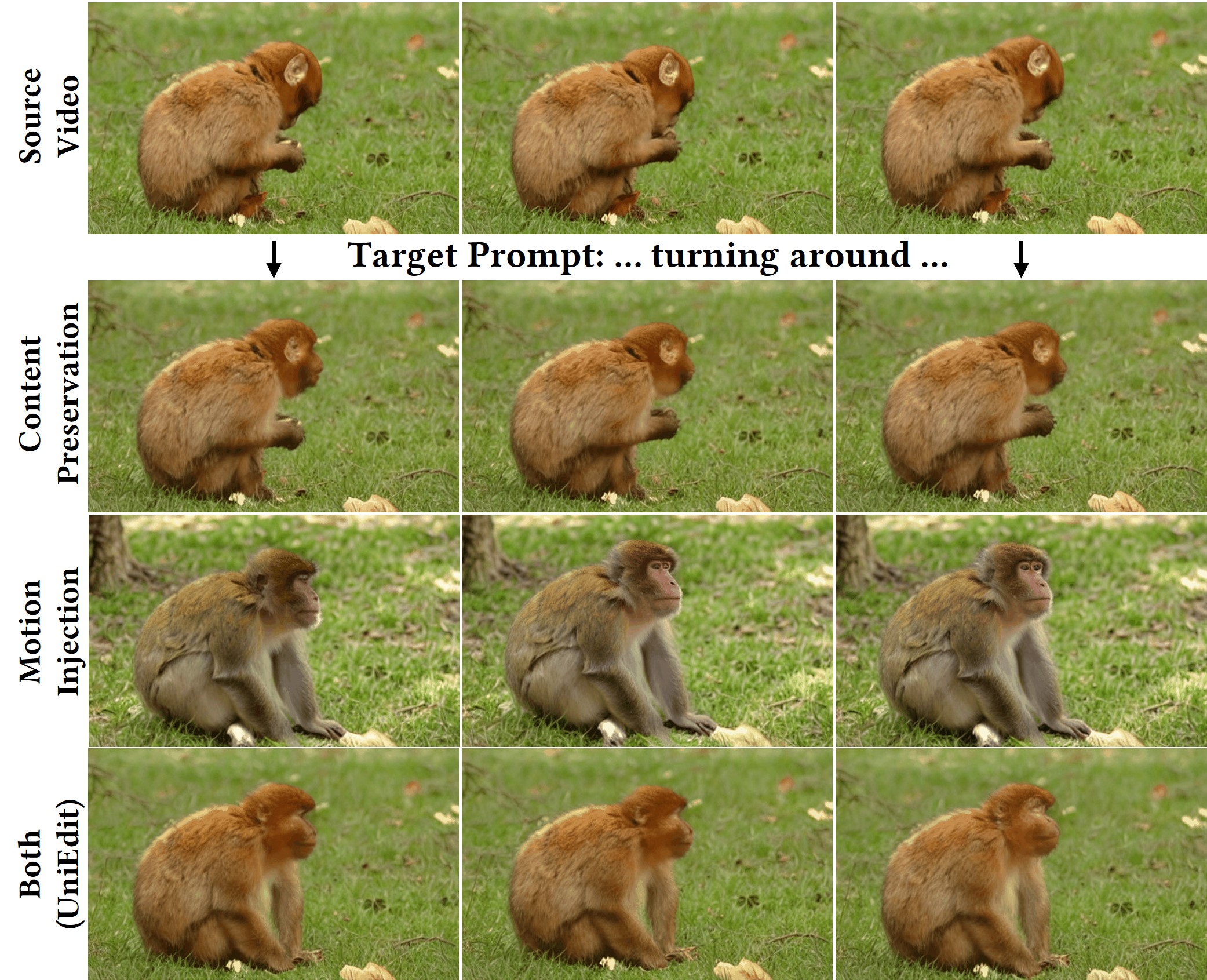}
\vspace{-6mm}
\caption{The proposed content preservation and motion injection both contribute to the final results.}
\label{fig_ablation1}
\vspace{-2mm}
\end{figure}

\begin{figure}[!t]\centering
\includegraphics[width=0.48\textwidth]{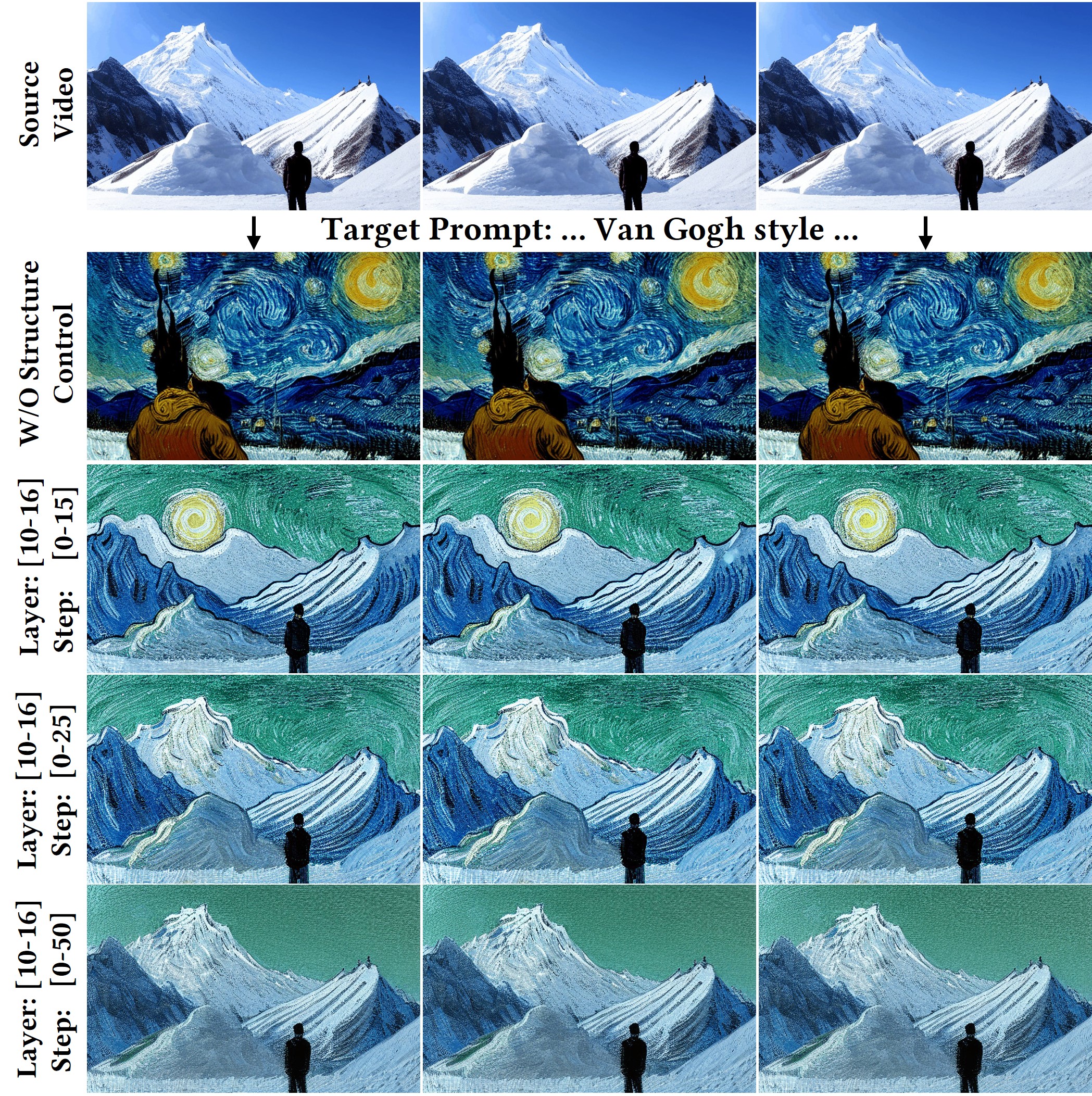}
\vspace{-6mm}
\caption{Ablation on spatial structure control and controllable video editing.}
\label{fig_ablation2}
\vspace{-3mm}
\end{figure}

\paragraph{How UniEdit Works?}
To better understand how UniEdit works and reveal our insight on the spatial and temporal self-attention layers, we visualize the features in SA-S and SA-T modules and compare them with the magnitude of optical flow between adjacent frames in Fig.~\ref{fig_vis}. It can be illustrated that, compared to the spatial query maps (second row), the temporal cross-frame attention maps (third row) show a significantly larger overlap with the optical flow (fourth row). Therefore, we leverage that the temporal self-attention layers encode the inter-frame dependency and conduct motion injection in temporal self-attention layers. We also conduct content preservation and structure control in spatial self-attention layers.

\vspace{-2mm}
\paragraph{The Two Auxiliary Branches Both Contribute to the Final Results.}
In UniEdit, we propose to preserve the content from the reconstruction branch and transfer motion from the motion-reference branch via spatial and temporal self-attention respectively. To better understand the impact of the two auxiliary branches, we visualize the edited results of utilizing each branch individually in Fig.~\ref{fig_ablation1}. When performing the content preservation from the reconstruction branch, though the identity and background are well-preserved, the synthesized frames deviate from the target prompt. On the other hand, only implementing the motion injection results in dramatic changes in the background. Meanwhile, UniEdit achieves satisfying performance in both content preservation and adherence to text.

\vspace{-2mm}
\paragraph{Spatial Structure Control Retains the Source Structure during the Appearance Editing.}
We conduct the ablation study on spatial structure control in the upper two rows of Fig.~\ref{fig_ablation2}, which demonstrates that the source structure is lost without the used structure control scheme, while our results (the lower three rows) adhere to the source structure.

\vspace{-2mm}
\paragraph{Controllable Video Editing.}
Our UniEdit supports controllable video editing by injecting the features into the main editing path at different layers and denoising steps. As exemplified in Fig.~\ref{fig_ablation2}, compared with injection at more steps ($t_2$=50), injection at fewer ($t_2$=15) steps produces a more stylized output, enabling flexible video editing.

\vspace{-2mm}
\section{Conclusion and Limitations}
In this paper, we design a novel tuning-free framework UniEdit for both video motion and appearance editing. By leveraging an auxiliary motion-reference branch and an auxiliary reconstruction branch and injecting features into the main editing path, it is capable of performing motion editing and various appearance editing.
There are nevertheless some limitations. Firstly, though editing on both motion and appearance can be achieved by calling UniEdit twice, it remains to be explored how to perform both types of editing simultaneously. Secondly, since we have multiple hyper-parameters, it is worth investigating the automatic determination scheme. We hope our work can promote the research toward unified content editing.

\vspace{-2mm}
\paragraph{Ethical Consideration.}
UniEdit is a tuning-free approach and is developed for research purposes only. It is suggested to add watermarks to prevent misuse.

\clearpage
\bibliography{example_paper}
\bibliographystyle{icml2024}

\newpage
\appendix
\onecolumn
We organize the Appendix as follows: 

$\bullet$ Appendix~\ref{appendix_sec_B}: detailed descriptions of experimental settings.

$\bullet$ Appendix~\ref{appendix_sec_C}: more experimental results, including ablation study on mask-guided coordination in (Section \ref{appendix_sec_C1}), more comparisons with baseline methods (Section \ref{appendix_sec_C2}), and more editing results of UniEdit (Section \ref{appendix_sec_C3}).

\section{Detailed Experimental Settings}
\label{appendix_sec_B}

\paragraph{Base T2V Model.}
We instantiate the proposed method on LaVie~\cite{lavie}, which is a pre-trained text-to-video generation model that produces consistent and high-quality videos. To achieve a fair comparison, we only leverage the base T2V model in \href{https://github.com/Vchitect/LaVie}{LaVie} and load the open-source pre-trained weights for video editing tasks in the experiments. Note that the edited video clip could further be seamlessly fed into the temporal interpolation model and the video super-resolution model to obtain video with a longer duration and higher resolution.

\paragraph{Video Preprocessing.}
For each input video, we resize it to the resolution of $320 \times 512$, followed by normalization, which is consistent with the training configuration of LaVie. Then, the pre-processed video is fed into the base model of Lavie to perform video editing. To maximize the generation power of LaVie, we set all input videos to 16 frames. For a source video, it takes 1-2 minutes to edit on an NVIDIA A100 GPU.

\paragraph{Configurations.}
For real source videos, we inverse them with $50$ DDIM inversion steps and perform DDIM deterministic sampling with $50$ steps for generation. 
For the generated videos, we use the same start latent of synthesizing the source video as the initial noise $z_T$ for the main editing path and two auxiliary branches. We use the commonly used classifier-free guidance technique \cite{ho2022classifier} with a scale of $7.5$.

\paragraph{Details of User Study.} 
As a text-guided editing task, in addition to CLIP scores, it is crucial to evaluate results through human subjective assessment. To achieve this, we utilized MOS (Mean Opinion Score) as our metric and collected feedback from $10$ experienced volunteers. We randomly selected $20$ editing samples and permuted results from different models. Volunteers were then tasked to evaluate the results based on two perspectives: frame consistency and textual alignment. They provided ratings for these aspects on a scale of $1$-$5$. Specifically, frame consistency measures the smoothness of the video, aiming to avoid dramatic jittering and ensure coherence between the content of each frame. Textual alignment assesses whether the editing results adhere to the text guidance and maintain the content of the source video. In the end, we computed the average user ratings for each method as our final results.
 
As illustrated in Tab.~\ref{tab_quality_eval}, UniEdit shows the best performance on frame consistency. Regarding textual alignment, UniEdit significantly outperforms all other baselines, demonstrating its capacity to support diverse editing scenarios.

\paragraph{Baselines.}
We implement all baseline methods with their official repositories. For MasaCtrl \cite{masactrl}, we adapt it to video editing by first setting the base model to a T2V model \cite{lavie}, then performing MasaCtrl on all frames of the source video. Moreover, since most baselines use StableDiffusion (SD) as the base model, we resize the source video to $512 \times 512$ to align with the default configuration of SD, then feed it into the denoising model, which can maximize the power of SD.

\section{Additional Experimental Results and Analysis}
\label{appendix_sec_C}

\subsection{More Ablation Study}
\label{appendix_sec_C1}

\begin{figure*}[t]\centering
\includegraphics[width=1\textwidth]{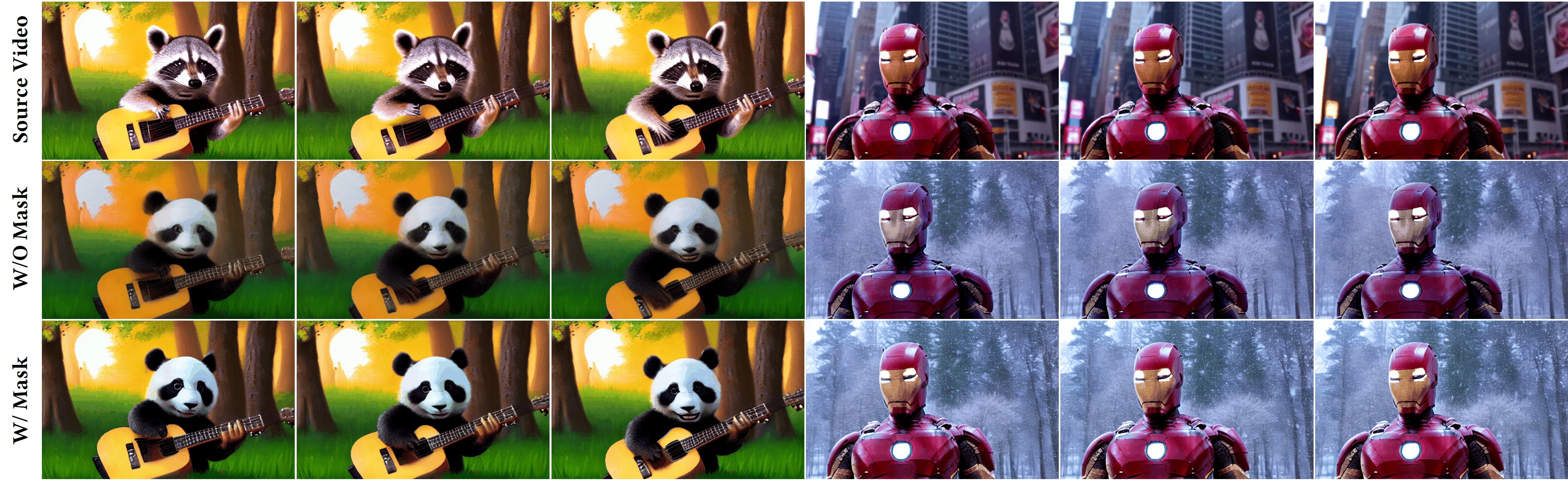}
\caption{Ablation on mask-guided coordination.}
\label{fig_appendix_ablation}
\end{figure*}

Please refer to Fig. \ref{fig_appendix_ablation} for the ablation on mask-guided coordination, where the second line indicates the results without using mask-guided coordination and the last line is UniEdit results with the proposed mask-guided coordination. It's observed that mask-guided coordination could further improve the editing performance.

\subsection{More Comparison with State-of-the-Art Methods}

\begin{figure*}[t]\centering
\includegraphics[width=1\textwidth]{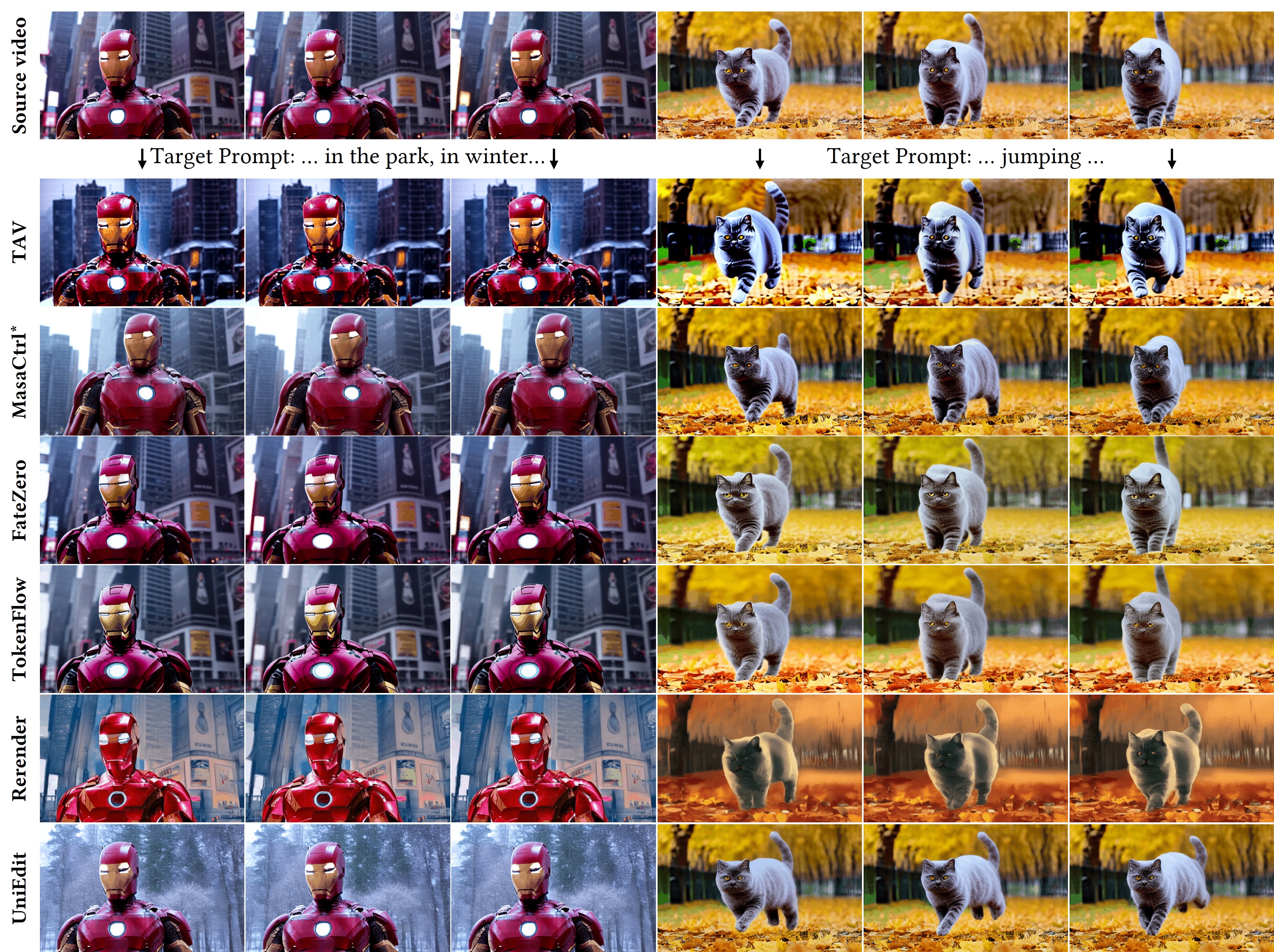}
\caption{More comparison with state-of-the-art methods.}
\label{fig_appendix_compare1}
\end{figure*}

\begin{figure*}[t]\centering
\includegraphics[width=1\textwidth]{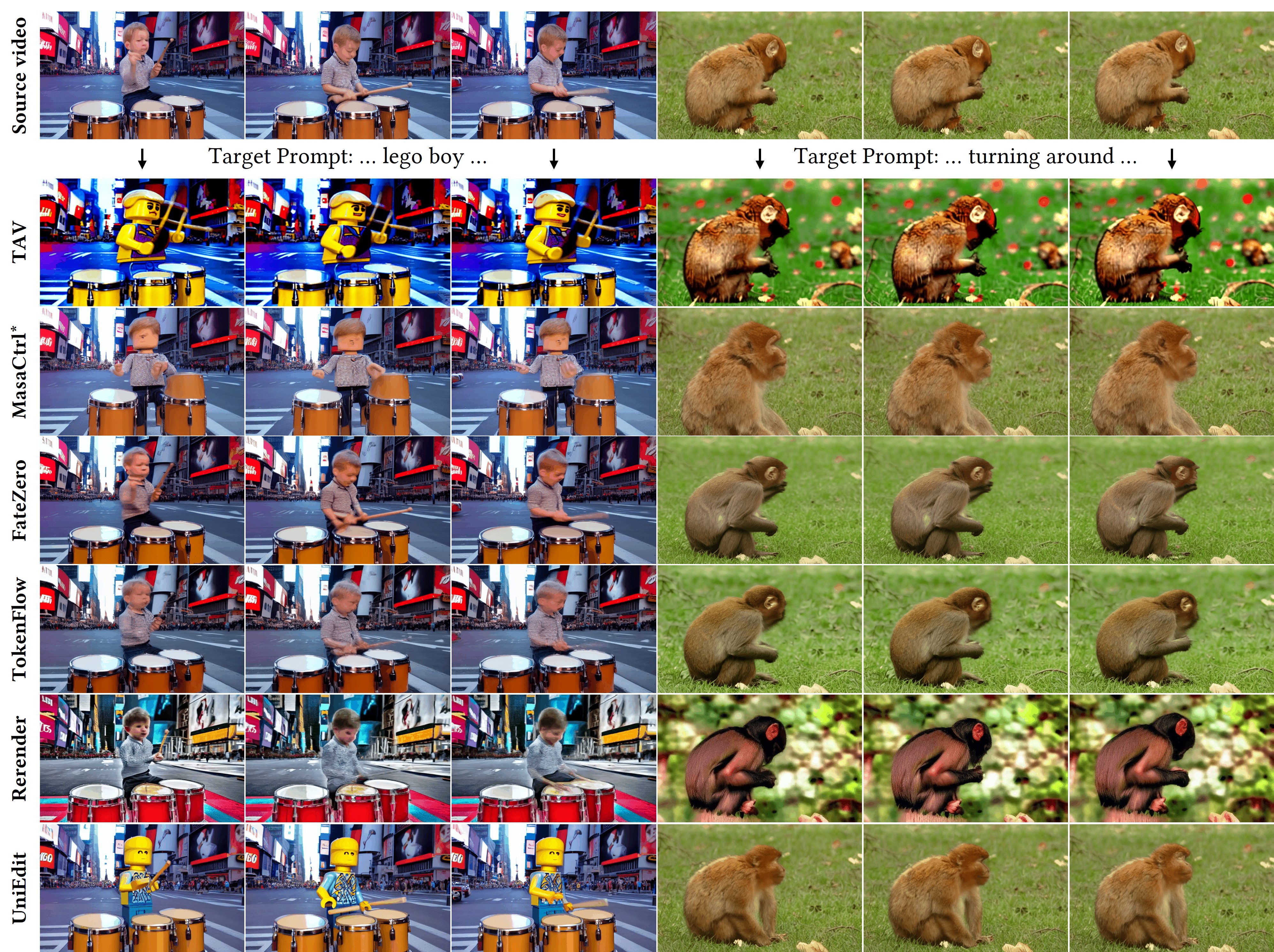}
\caption{More comparison with state-of-the-art methods.}
\label{fig_appendix_compare2}
\end{figure*}

Please refer to Fig. \ref{fig_appendix_compare1} and Fig. \ref{fig_appendix_compare2} for more comparison with the state-of-the-art methods.

\label{appendix_sec_C2}
\subsection{More Results of UniEdit}
\label{appendix_sec_C3}

\begin{figure*}[!h]\centering
\includegraphics[width=1\textwidth]{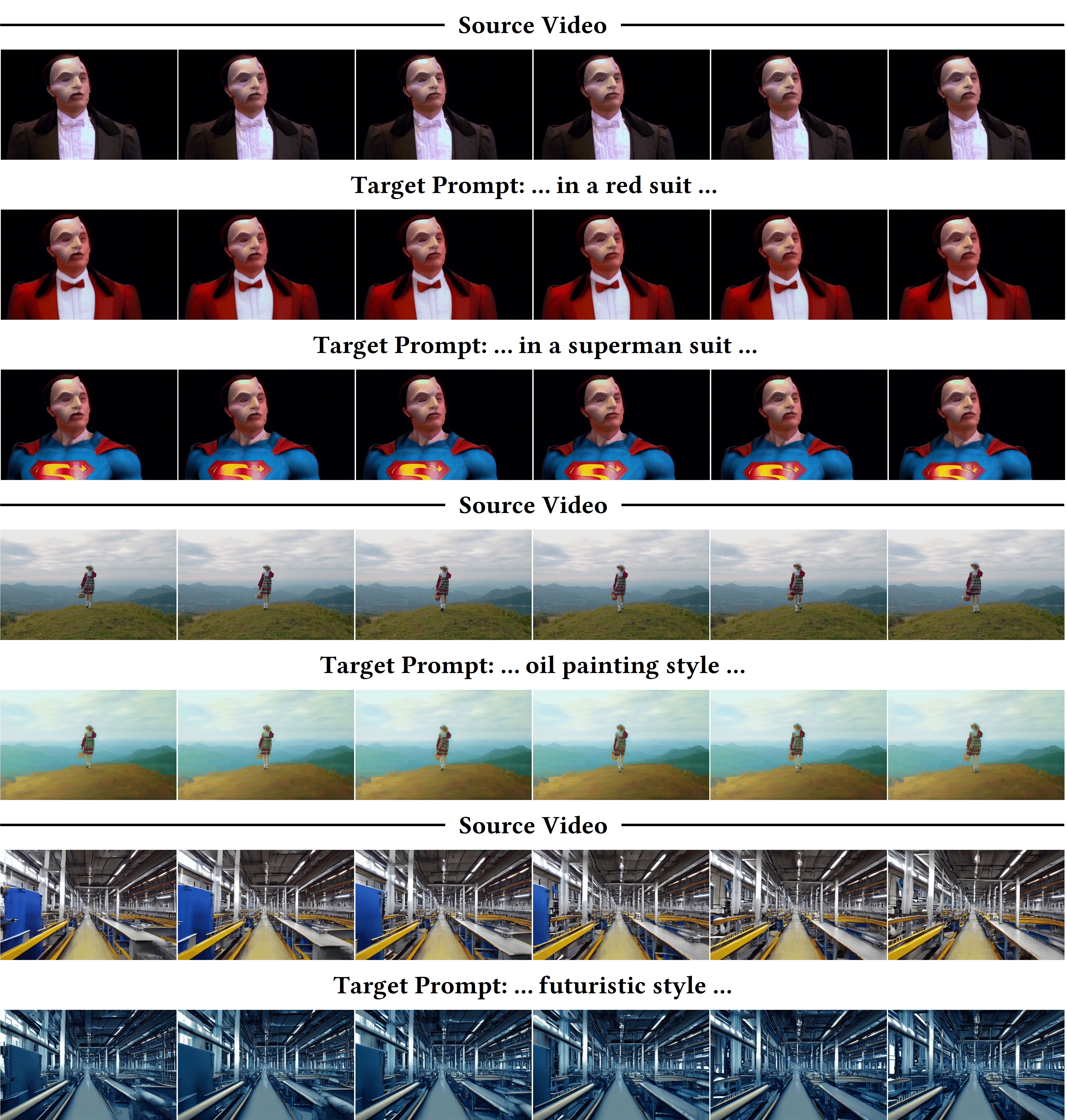}
\caption{More appearance editing results of UniEdit.}
\label{fig_appendix_ours_appear1}
\end{figure*}

\begin{figure*}[t]\centering
\includegraphics[width=1\textwidth]{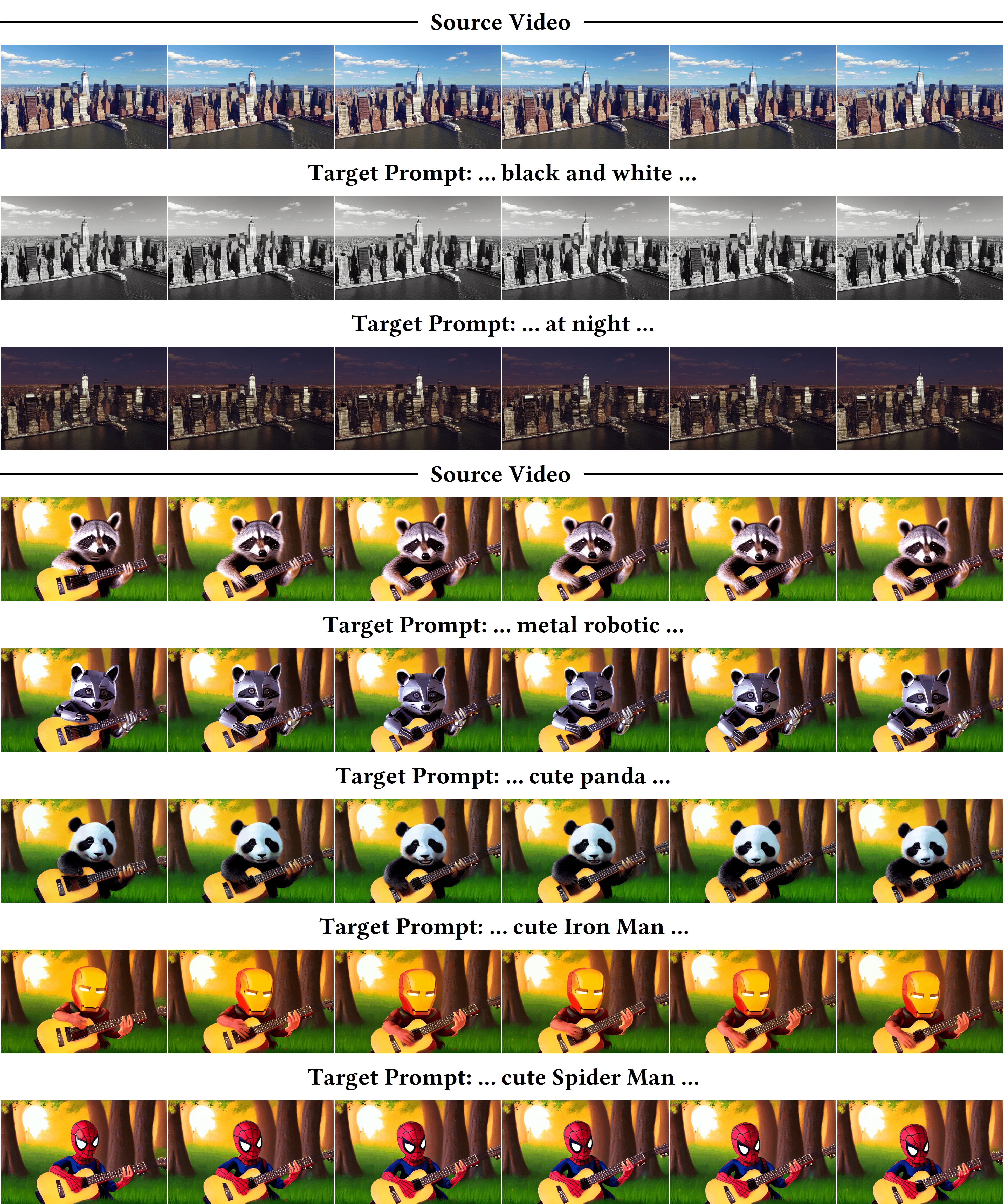}
\caption{More appearance editing results of UniEdit.}
\label{fig_appendix_ours_appear2}
\end{figure*}

\begin{figure*}[t]\centering
\includegraphics[width=1\textwidth]{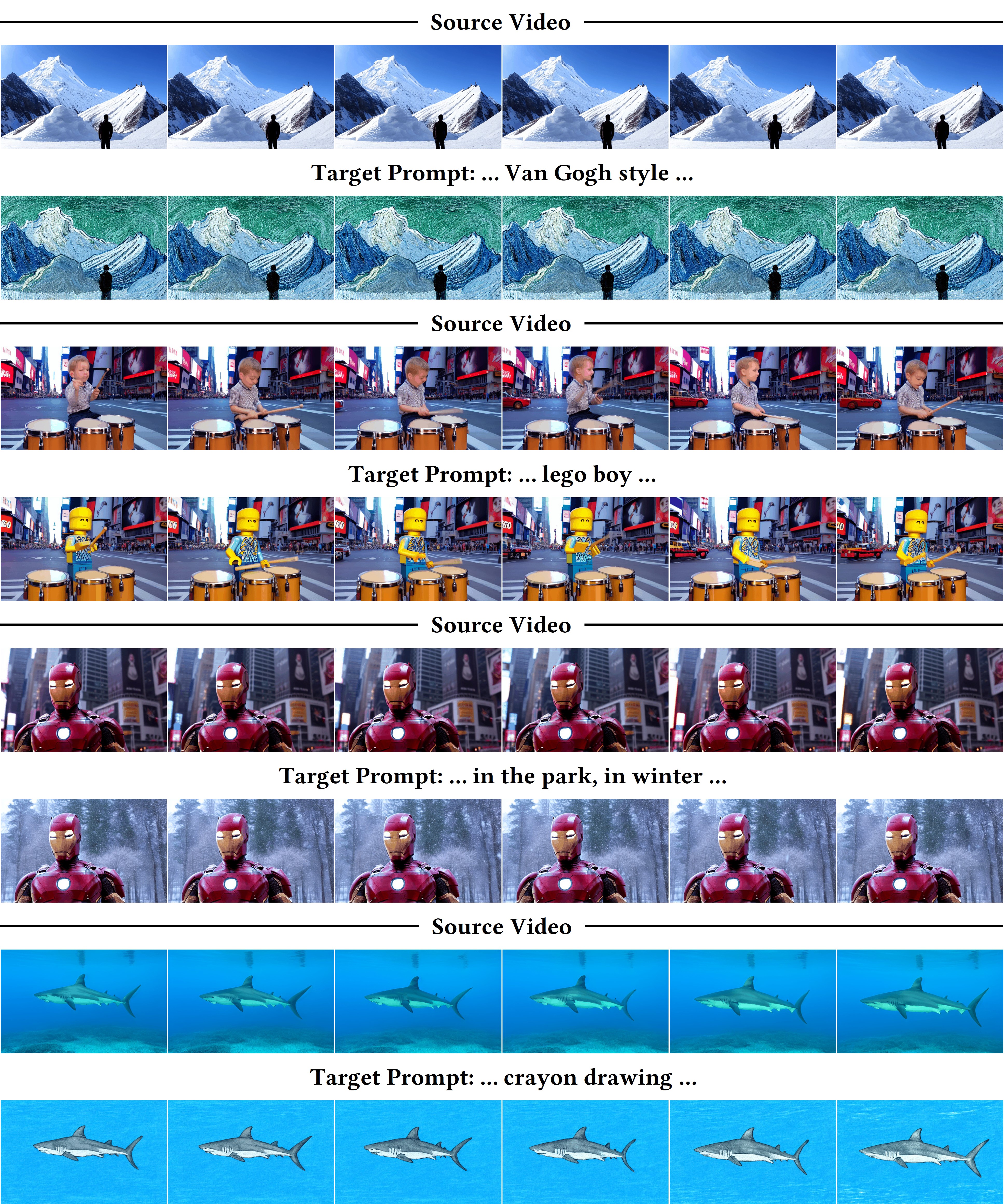}
\caption{More appearance editing results of UniEdit.}
\label{fig_appendix_ours_appear3}
\end{figure*}

\begin{figure*}[t]\centering
\includegraphics[width=1\textwidth]{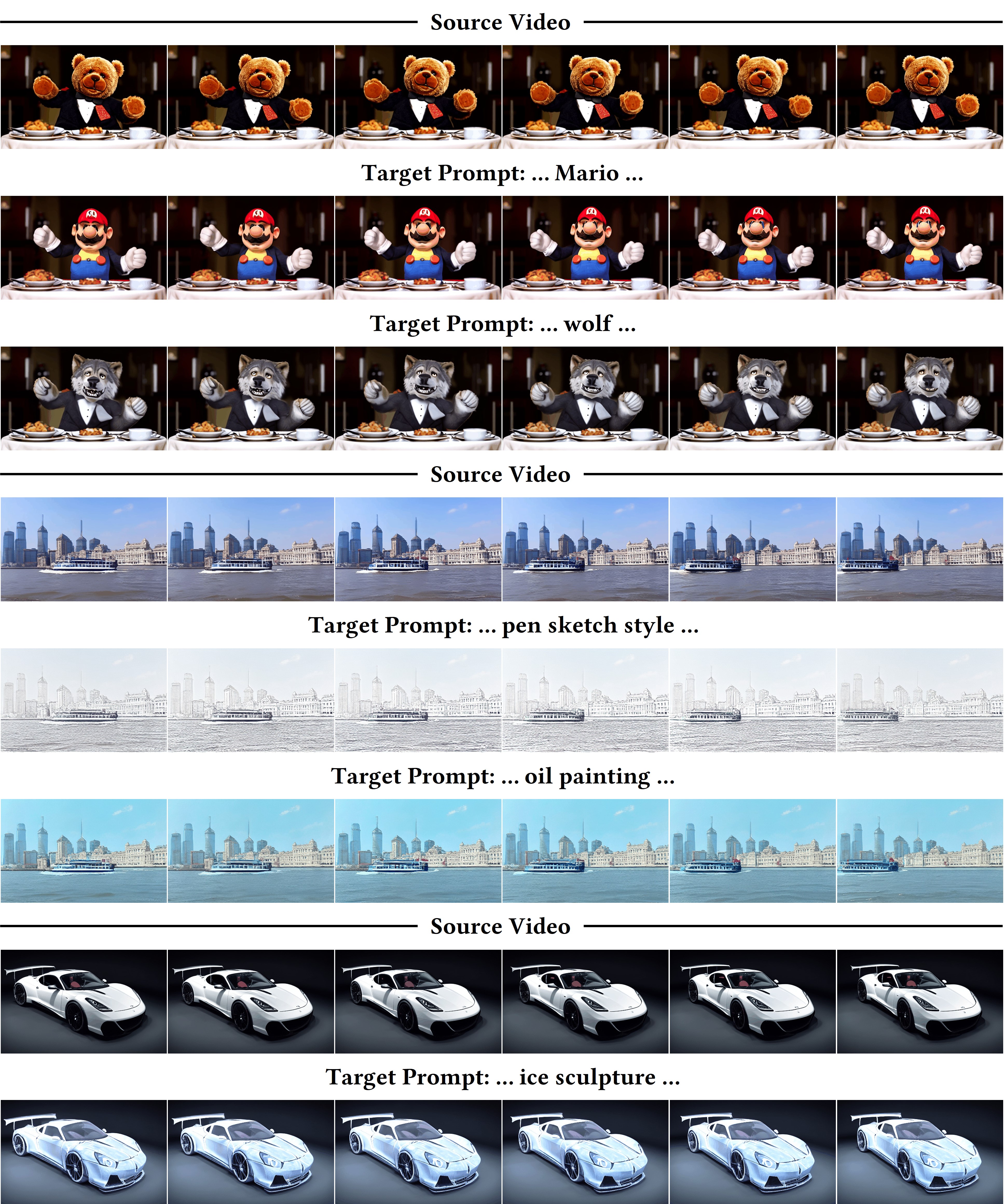}
\caption{More appearance editing results of UniEdit.}
\label{fig_appendix_ours_appear4}
\end{figure*}

\begin{figure*}[t]\centering
\includegraphics[width=1\textwidth]{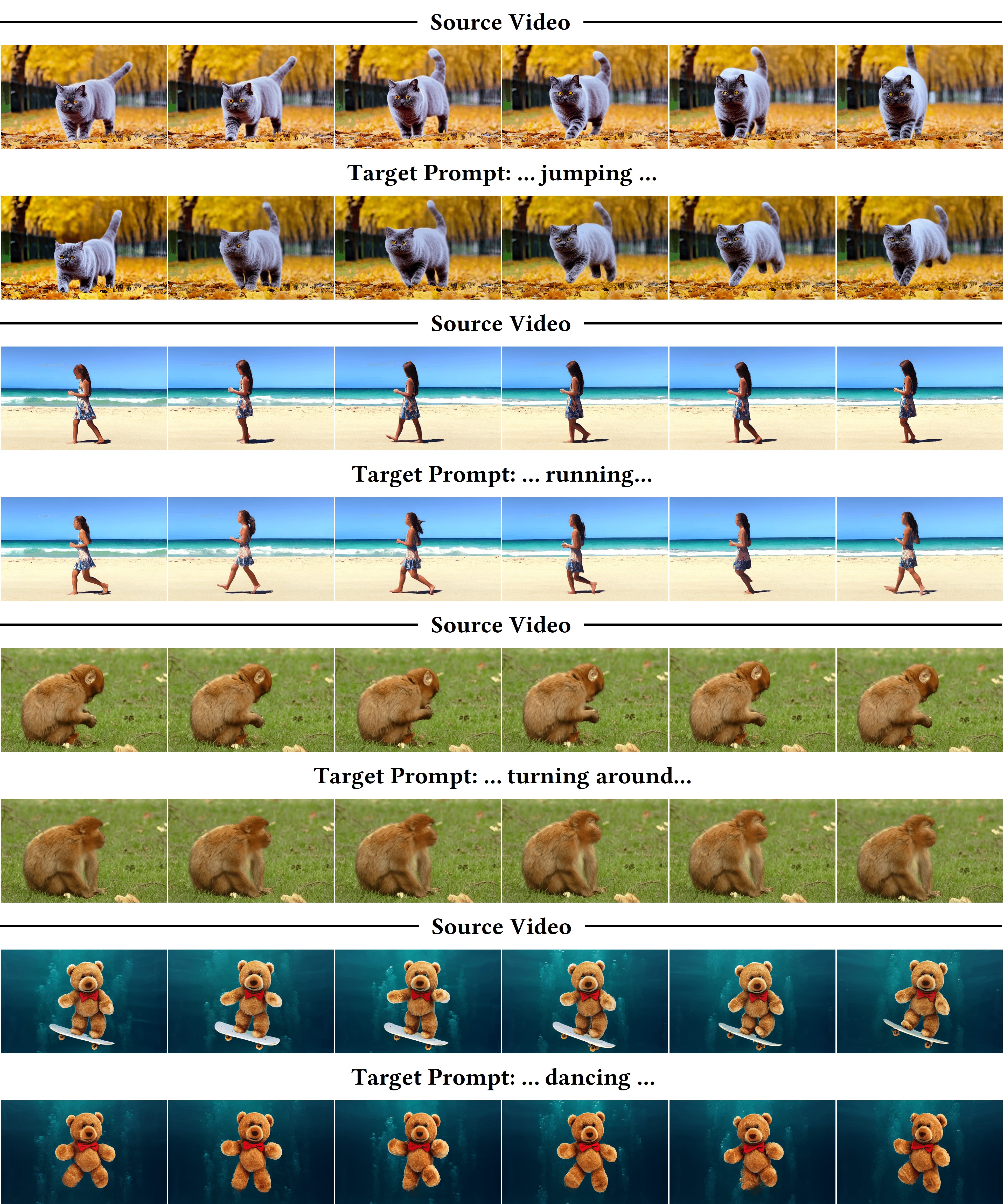}
\caption{More motion editing results of UniEdit.}
\label{fig_appendix_ours_motion1}
\end{figure*}

\begin{figure*}[t]\centering
\includegraphics[width=1\textwidth]{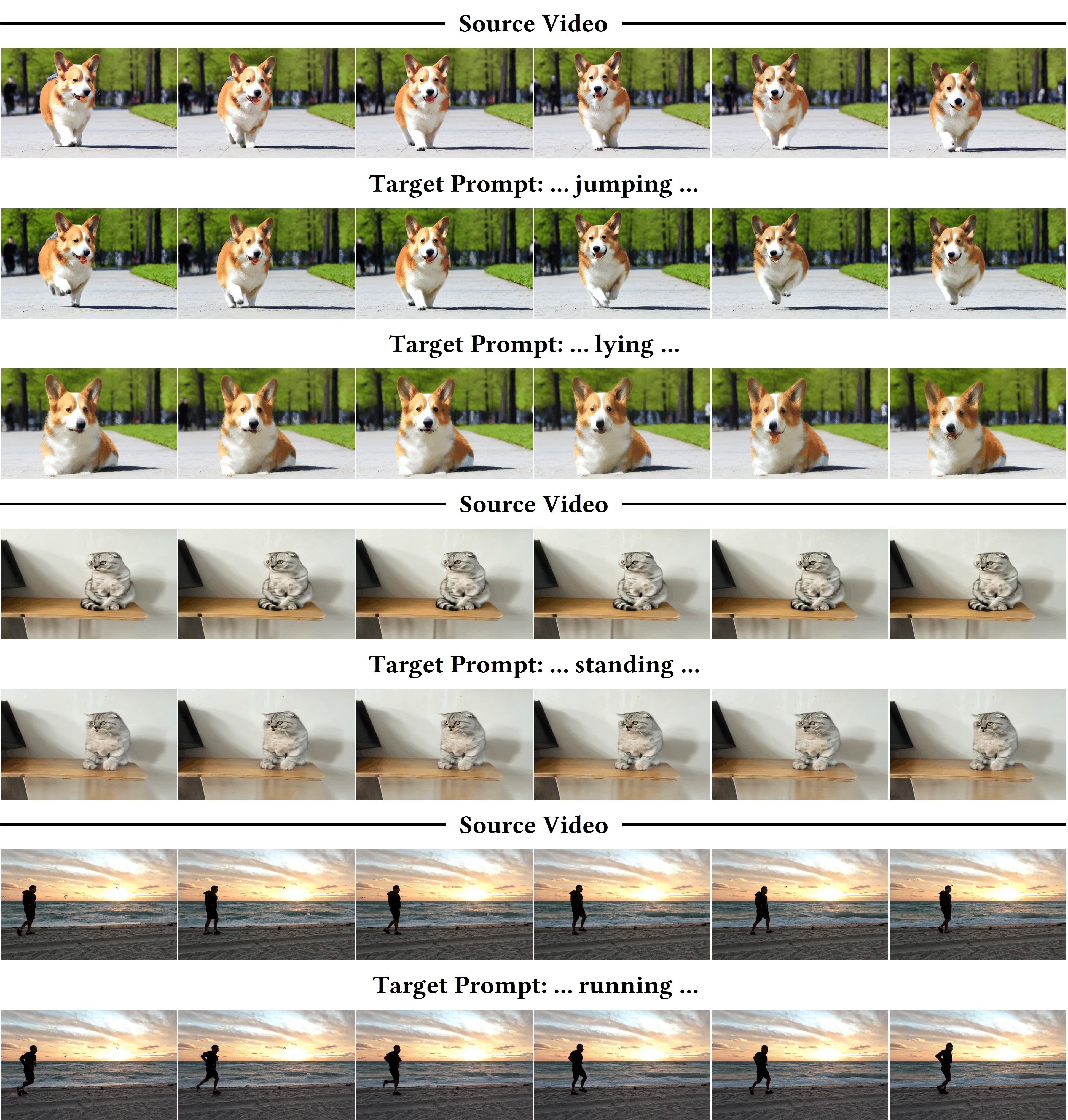}
\caption{More motion editing results of UniEdit.}
\label{fig_appendix_ours_motion2}
\end{figure*}

\begin{figure*}[t]
\includegraphics[width=1\textwidth]{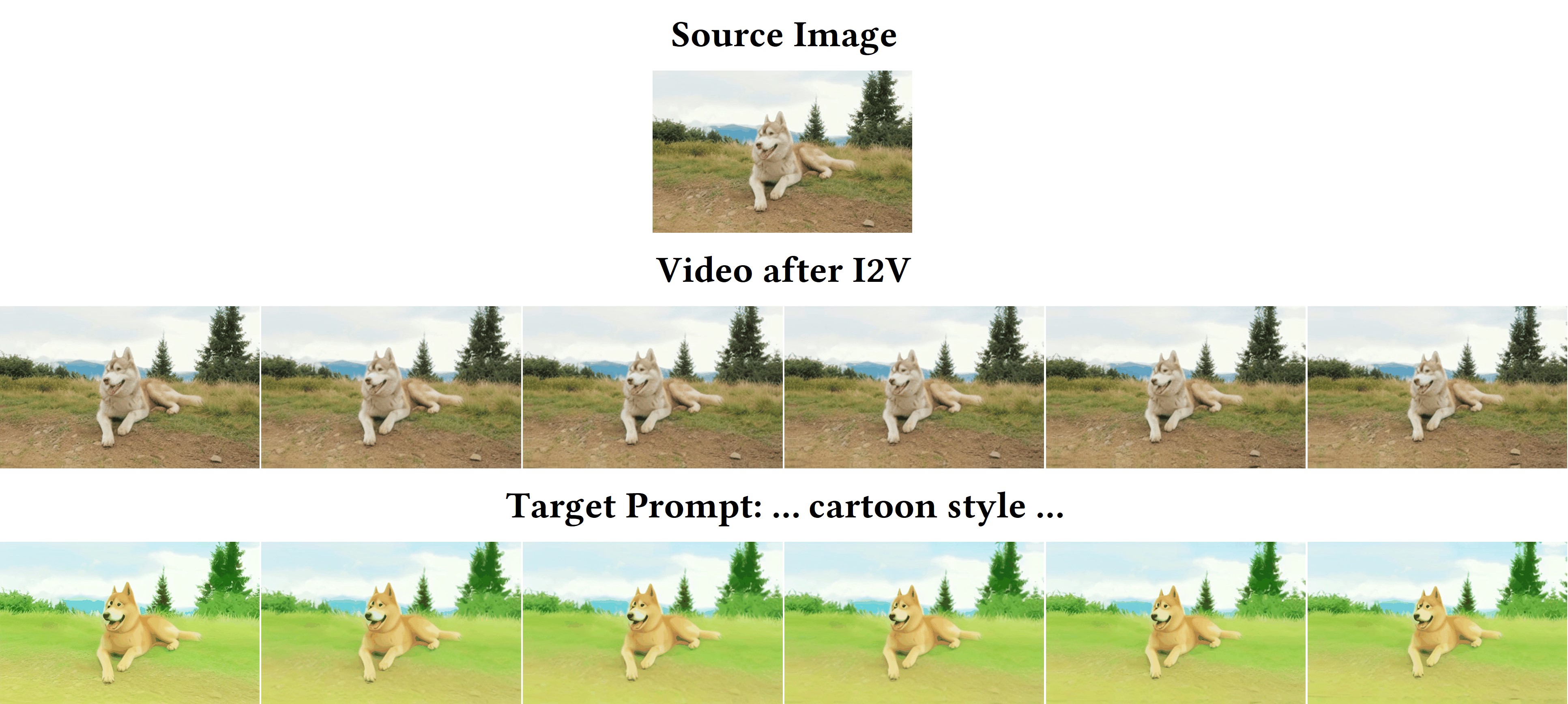}
\caption{Results of text-image-to-video synthesis in Sec. \ref{sec_ti2v}.}
\label{fig_appendix_ours_ti2v}
\end{figure*}

More edited results of UniEdit are provided in Fig.~\ref{fig_appendix_ours_appear1}-\ref{fig_appendix_ours_motion2}. Examples of TI2V generation are provided in Fig.~\ref{fig_appendix_ours_ti2v}.

\clearpage

\clearpage

\clearpage

\clearpage

\clearpage

\clearpage

\end{document}